\documentclass[runningheads]{llncs}
\usepackage{graphicx}
\usepackage{tikz}
\usepackage{comment}
\usepackage{amsmath,amssymb} % define this before the line numbering.
\usepackage{color}
\usepackage{subfig}
\usepackage{booktabs}
\usepackage{multirow}
\usepackage{multicol}
\usepackage{rotating}
\usepackage[accsupp]{axessibility}  % Improves PDF readability for those with disabilities.

\usepackage{xspace}
\makeatletter
\DeclareRobustCommand\onedot{\futurelet\@let@token\@onedot}
\def\@onedot{\ifx\@let@token.\else.\null\fi\xspace}

\def\eg{\emph{e.g}\onedot} 
\def\ie{\emph{i.e}\onedot}

\def\etal{\emph{et al}\onedot}
\makeatother

\newcommand{\PseudoClick}{\texttt{PseudoClick}}
\DeclareMathOperator{\NFL}{L}

\usepackage[accsupp]{axessibility}  % Improves PDF readability for those with disabilities.

\begin{document}
% \linenumbers
\pagestyle{headings}
\mainmatter
\def\ECCVSubNumber{2903}  % Insert your submission number here

\title{PseudoClick: Interactive Image Segmentation with Click Imitation} % Replace with your title

%\begin{comment}
% \titlerunning{ECCV-22 submission ID \ECCVSubNumber} 
% \authorrunning{ECCV-22 submission ID \ECCVSubNumber} 
% \author{Anonymous ECCV submission}
% \institute{Paper ID \ECCVSubNumber}
%\end{comment}
%******************

% \begin{comment}
\titlerunning{PseudoClick: Interactive Image Segmentation with Click Imitation}

% \author{
% Qin Liu\inst{1,2}\orcidlink{0000-0001-6342-5311} \and
% Meng Zheng\inst{2}\orcidlink{0000-0002-6677-2017} \and \\
% Benjamin Planche\inst{2}\orcidlink{0000-0002-6110-6437} \and
% Srikrishna Karanam\inst{2}\orcidlink{0000-0002-7627-7765}  \and 
% Terrence Chen\inst{2} \and % TODO
% Marc Niethammer\inst{1} \and % TODO
% Ziyan Wu\inst{2}\orcidlink{0000-0002-9774-7770}}

\author{
Qin Liu\inst{1,2} \and
Meng Zheng\inst{2} \and \\
Benjamin Planche\inst{2} \and
Srikrishna Karanam\inst{2} \and 
Terrence Chen\inst{2} \and % TODO
Marc Niethammer\inst{1} \and % TODO
Ziyan Wu\inst{2}}

\authorrunning{Q. Liu et al.}
\institute{
University of North Carolina at Chapel Hill, Chapel Hill NC, USA \and
United Imaging Intelligence, Cambridge MA, USA \\
\email{\{first.last\}@uii-ai.com}, \email{qinliu19@cs.unc.edu}}
% \end{comment}
%******************
\maketitle

\begin{abstract}
The goal of click-based interactive image segmentation is to obtain precise object segmentation masks with limited user interaction, \ie, by a minimal number of user clicks. 
Existing methods require users to provide all the clicks: by first inspecting the segmentation mask and then providing points on mislabeled regions, iteratively. 
We ask the question: can our model directly predict where to click, so as to further reduce the user interaction cost?
To this end, we propose {\PseudoClick}, a generic framework that enables existing segmentation networks to propose candidate next clicks. These automatically generated clicks, termed pseudo clicks in this work, serve as an imitation of human clicks to refine the segmentation mask.
We build {\PseudoClick} on existing segmentation backbones and show how the click prediction mechanism leads to improved performance.
We evaluate {\PseudoClick} on 10 public datasets from different domains and modalities, showing that our model not only outperforms existing approaches but also demonstrates strong generalization capability in cross-domain evaluation. We obtain new state-of-the-art results on several popular benchmarks, \eg, on the Pascal dataset, our model significantly outperforms existing state-of-the-art by reducing 12.4\% number of clicks to achieve 85\% IoU.
\end{abstract}

% \bnote{Are we planning to add more content? Otherwise, we could convert some '\\noindent\\textbf{}' headlines into actual subsection/paragraph headlines, to increase the number of lines.}

\section{Introduction}
\label{sec:intro}
Recent years have seen tremendous progresses in segmentation methods for various  applications, \eg, semantic object/instance segmentation~\cite{garcia2017review,minaee2021image}, video understanding~\cite{yang2019video,xu2018youtube}, autonomous driving~\cite{cordts2016cityscapes,geiger2012we,neuhold2017mapillary}, and medical image analysis~\cite{shen2017deep,litjens2017survey}. The success of these applications heavily relies on the availability of large-scale pixel-level annotation masks, which are very laborious and costly to obtain. Interactive image segmentation, which aims at extracting the object-of-interest using limited human interactions, is an efficient way to obtain these annotations. 
% Hence, significant research efforts are ongoing to explore interactive segmentation approaches.

\begin{figure}
  \centering
  \includegraphics[width=7.2cm, height=4.2cm]{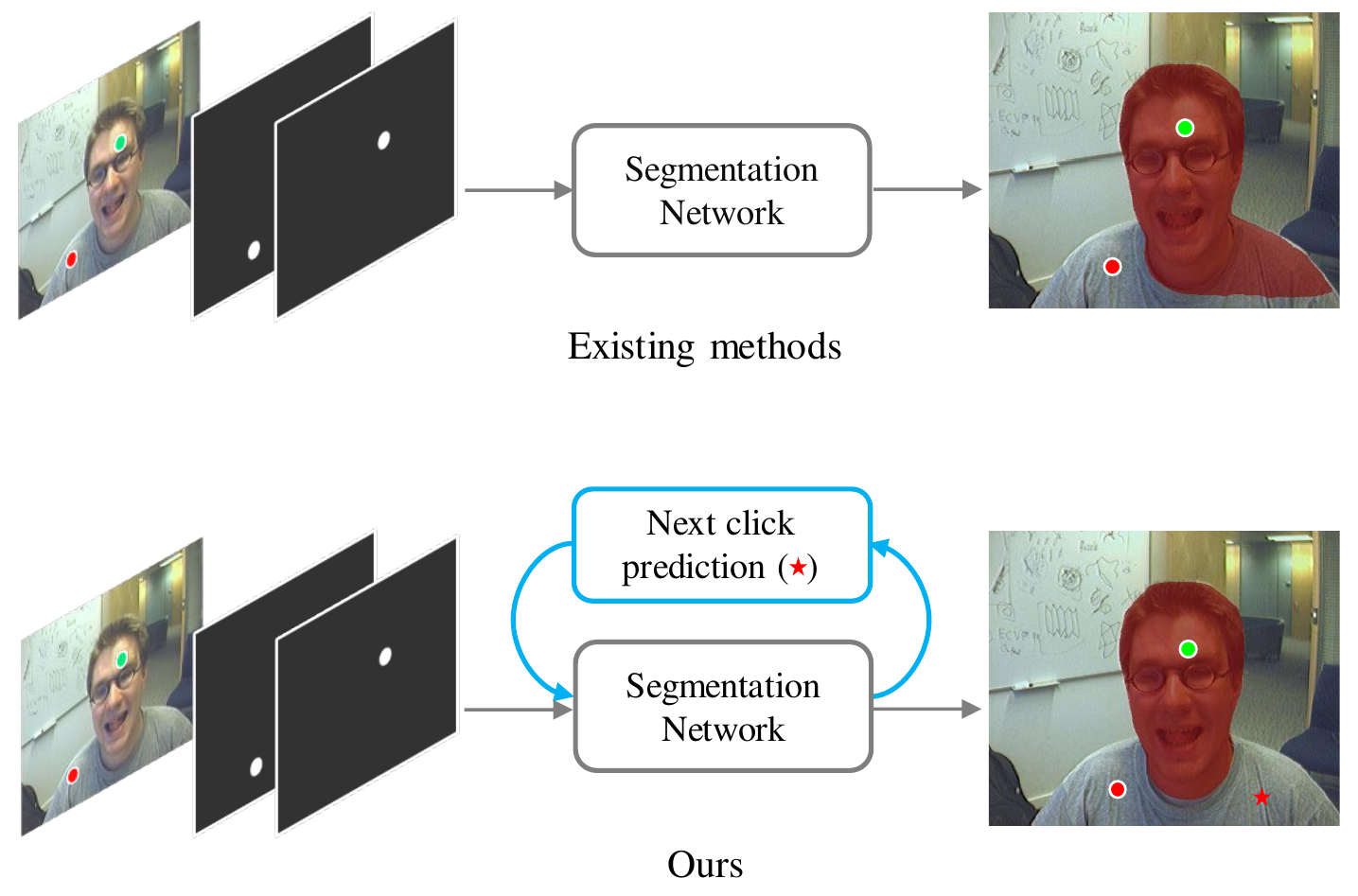}
  \caption{The key difference between existing methods and our method is the ability for click prediction. (a) Illustration of existing methods. All clicks are provided by users. (b) Illustration of our method, which predicts the next click based on the current segmentation mask. This predicted pseudo click is used to refine the segmentation mask. Note the manual segmentation is not required at test time for the pseudo-click prediction.}
  \label{fig:pc_sketch}
\end{figure}

While different interaction types have been investigated, including clicks \cite{xu2016deep,sofiiuk2021reviving}, bounding boxes \cite{xu2017deep,zhang2020interactive,wu2014milcut}, and scribbles~\cite{boykov2001interactive,rother2004grabcut}, we only focus on click-based interactive segmentation because it has the simplest interaction, and well-established training and evaluation protocols~\cite{xu2016deep,sofiiuk2021reviving}. Compared with scribble-based methods~\cite{boykov2001interactive,rother2004grabcut}, click-based interactive segmentation requires no heuristics or complex procedures to simulate user input.
Recent work on click-based interactive segmentation has resulted in state-of-the-art segmentation performance using various inference-time optimization schemes~\cite{sofiiuk2020f}, which are computationally expensive due to multiple backward passes during inference. More recently, Sofiiuk \etal~proposed RITM~\cite{sofiiuk2021reviving}, a simple feedforward approach requiring no inference-time optimization for click-based interactive segmentation that shows performance superior to all existing models when trained on the combined COCO~\cite{lin2014microsoft} and LVIS~\cite{gupta2019lvis} datasets with diverse and high-quality annotations.

While existing methods in this area have shown improved performance, \eg, using inference-time optimization schemes~\cite{sofiiuk2020f} or iterative mask-guided training schemes~\cite{sofiiuk2021reviving}, these methods still require users to provide all the clicks. Hence, users need to interactively inspect the resulting segmentation masks and then provide points for mislabeled regions. We ask the question: can the segmentation model directly predict where to click so as to further reduce the number of human clicks? To this end, we propose {\PseudoClick}, a novel framework for interactive segmentation that equips existing segmentation backbones with the ability to predict user clicks automatically. As shown in Fig.~\ref{fig:pc_sketch}, the key difference between existing methods and our proposed one is the ability to generate additional, beneficial, and ``free" pseudo clicks as the prediction of human clicks for refining the segmentation mask.

{\PseudoClick} is a generic framework that can be built upon existing segmentation backbones, including both CNNs and transformers. We equip an existing segmentation backbone with clicks prediction mechanism in the following procedures: we first introduce an error decoder, in parallel with the segmentation decoder of the backbone, that produces the false-positive (FP) and false-negative (FN) error maps for the current segmentation mask. We then extract a pseudo click from either the FP or the FN map, depending on which map contains the largest error. Specifically, a positive click should be generated from the FN map, while a negative click should be generated from the FP map. After that, the generated pseudo click (we generate one pseudo click each time) will be updated in the network input to refine the segmentation mask for the next forward pass. Note that the entire process is an imitation of the core human activity in interactive segmentation: visually estimating the segmentation errors, \ie, over-segmentation (FP) or under-segmentation (FN), before determining what and where the next click should be. 

We evaluate our method extensively on \textbf{10} public datasets (see Sec. \ref{sec:evaluation_details}). 
Evaluation  results  show  that  our  model  not only outperforms existing approaches but also demonstrates strong generalization capabilitity in cross-domain evaluation on medical images. On the Pascal dataset, our model significantly outperforms existing state-of-the-art by reducing 12.4\% and 11.4\% number of clicks to achieve 85\% and 90\% IoU, respectively. For the cross-domain evaluation on BraTS~\cite{baid2021rsna} and ssTEM~\cite{gerhard2013segmented}, our method significantly outperforms existing approaches to a large margin. 
Our main contributions are: 

\begin{itemize}
    \item[1)] We propose {\PseudoClick}, a novel interactive segmentation framework that directly imitates human clicks through the segmentation network and refines the segmentation mask with these imitated pseudo clicks. Our proposed framework differs from existing interactive segmentation methods in that it provides additional, beneficial, and ``free" clicks during the annotation process for a human-in-the-loop.
    \item[2)] We show that {\PseudoClick} is an efficient and generic framework that can be built upon different types of segmentation backbones, including both CNNs and transformers, with little effort in tuning the hyper-parameters and modifying the network architectures.
    \item[3)] We evaluate {\PseudoClick} thoroughly on benchmarks from multiple domains and modalities with extensive comparison and cross-domain evaluation experiments. The results show that {\PseudoClick} not only outperforms existing state-of-the-art approaches on in-domain benchmarks but also demonstrates strong generalization capability on cross-domain evaluation.
\end{itemize}

\section{Related Work}
\label{sec:related}

\subsubsection{Click-based interactive image segmentation.}  
Click-based interactive segmentation has the simplest interaction and well-established training and evaluation protocols~\cite{xu2016deep,sofiiuk2021reviving}.
% requiring no heuristics or complex procedures to simulate user input compared with other types interaction such as bounding boxes~\cite{xu2017deep,zhang2020interactive,wu2014milcut} and scribbles~\cite{boykov2001interactive,rother2004grabcut}.
% iSegFormer~\cite{liu2021isegformer} first proposes a transformer-based interactive segmentation method and applies it to medical images.
% \bnote{let's maybe avoid referencing this later work, it would undermine this submission.}
Xu \etal~\cite{xu2016deep} first apply CNNs for interactive segmentation and propose a click simulation strategy for training that has inspired many future works~\cite{sofiiuk2021reviving,liu2021isegformer,chen2022focalclick}.
Compared with previous click-based approaches~\cite{sofiiuk2021reviving,song2018seednet,chen2022focalclick}, {\PseudoClick} is unique because it is the first work that imitates human clicks and refines the segmentation with automatically generated pseudo clicks.

\subsubsection{Other types of interactive feedback.} Other than clicks, different types of user interaction have been explored in this field, including bounding boxes~\cite{zhang2020interactive}, scribbles~\cite{cheng2021modular}, and interactions from multiple modalities \cite{ding2020phraseclick}. 
The main drawback of bounding box-based approaches is the lack of a specific object reference inside the selected region, as well as a clear approach to correct the predicted mask. 
Inside outside guidance (IOG)~\cite{zhang2020interactive} addresses this issue by combining clicks with bounding points of the target object and by allowing corrections of the predicted mask. 
Our method differs from all these in that we only use clicks as the interaction. Instead of exploring more complicated interactions, we try to imitate user clicks and to decrease their number required to acquire predefined accuracy. 

\subsubsection{Imitation learning and beyond.}
Imitation learning aims at mimicking human behavior for a given task~\cite{hussein2017imitation}.
While this field has recently gained attention due to computing and sensing advances as well as the rising demand for intelligent applications~\cite{osa2018algorithmic,zhang2018deep}, it has never been explored in the interactive segmentation tasks. 
We claim that our method is highly related to imitation learning in that it imitates the core user activity in the interactive annotation process: visually estimating the segmentation errors before determining what and where the next point should be. 
SeedNet~\cite{song2018seednet} first proposes a reinforcement learning approach for automatically generating automatic click.
However, their method is not an imitation of human clicks because it automatically generates a sequence of clicks for achieving an implicit long-term reward without human intervention given the initial two clicks. Therefore, it is more like a post-processing approach.
In contrast, our method is an imitation of human clicks because it explicitly quantifies the FP\&FN errors and then generates the next click based on the estimated errors (just like a human annotator would do).
Our method imitates this process by introducing a pseudo click generation mechanism on existing segmentation backbones, as discussed in Sec.~\ref{sec:method}.
Our idea of predicting FP\&FN errors for segmentation is also related to the idea of ``prediction loss" proposed in \cite{yoo2019learning}, but the two ideas are investigated in significantly different problem settings.

\begin{figure*}
    \centering
    \includegraphics[width=11.0cm, height=5.5cm]{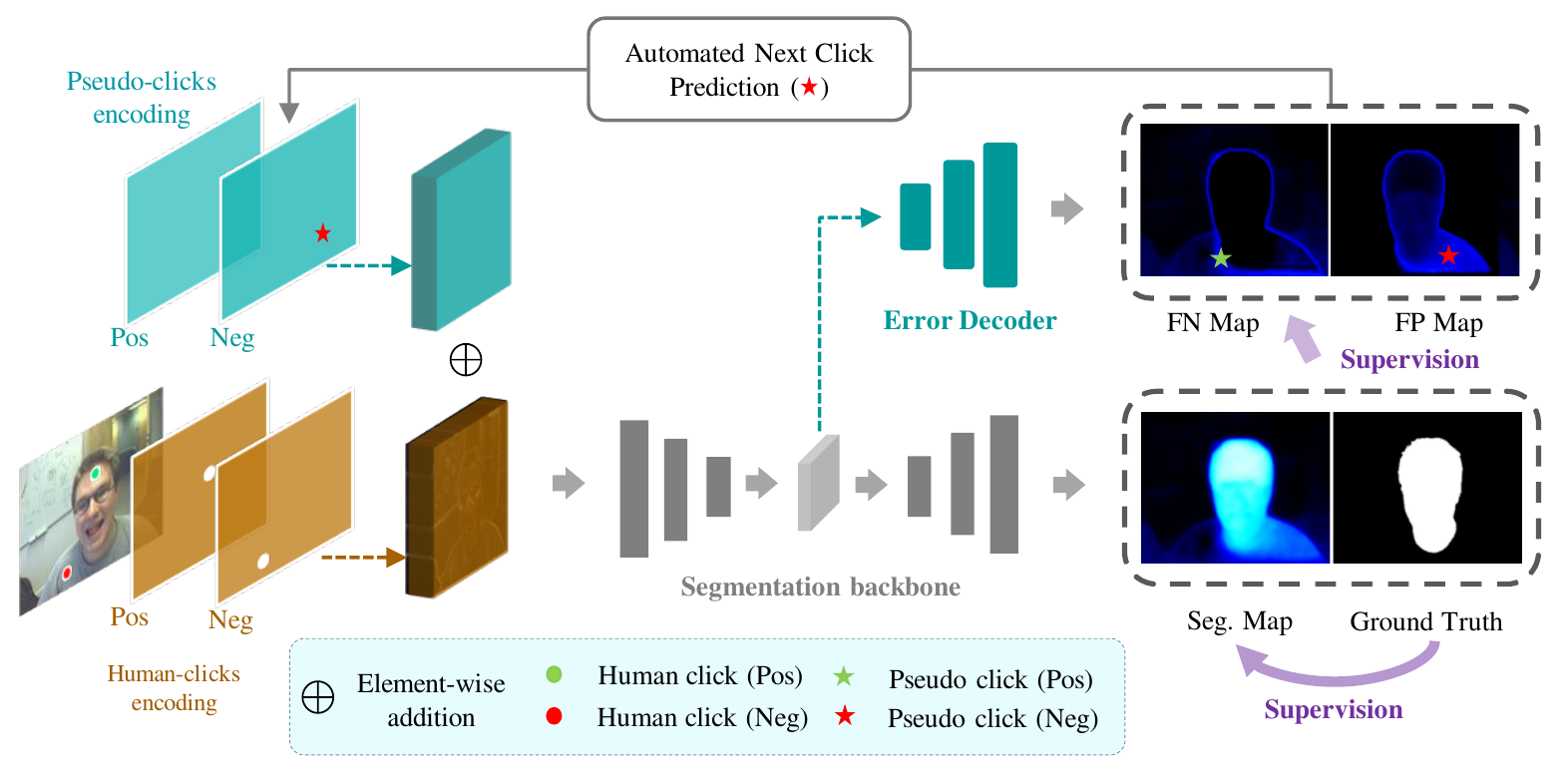}
   \caption{{\PseudoClick} overview. Given the image and clicks, the network outputs the segmentation map, along with two error maps that predict the false-positives and false-negatives of the segmentation mask, respectively. Then, a pseudo click will be generated from either the FP map or the FN map (see Sec.~\ref{sec:pseudo_clicks_generation}). After that, the network refines the segmentation mask in the second forward pass by adding the new pseudo click to the click encoding.}
   \label{fig:pc_framework}
\end{figure*}

\section{Method}
\label{sec:method}
As shown in Fig.~\ref{fig:pc_framework}, {\PseudoClick} is a generic framework that builds upon existing segmentation framework with two additional modules: a segmentation error decoder module and a clicks-encoding module that consists of encoding masks for both human clicks and pseudo clicks. In Sec.~\ref{sec:segmentation_error_decoder}, we first describe the segmentation error decoder that outputs two error maps from which pseudo clicks are generated. In Sec.~\ref{sec:pseudo_clicks_generation}, we further describe the pseudo clicks generation process. In Sec.~\ref{sec:pseudo_clicks_encoding}, we then introduce the encoding mechanism that transforms pseudo clicks into spatial signal for feeding into the segmentation backbone. In Sec.~\ref{sec:loss_function}, we introduce the loss function for training our models. In Sec.~\ref{sec:implementation_details}, we conclude the whole section by providing additional implementation details about the proposed method.

\subsection{Segmentation Error Decoder}
\label{sec:segmentation_error_decoder}

The segmentation error decoder is introduced in parallel with the segmentation decoder of the backbone. It generates two error maps that estimate the false positive (FP) and false negative (FN) errors of the segmentation mask. Both error maps are probability maps\footnote{Note that we call these \textit{probability maps}, but in general they will likely be miscalibrated. If desired, calibration can be improved, for example, using the approach in~\cite{ding2021local}.} that can be trained as two binary segmentation tasks.
The FP and FN maps generated by the error decoder are supervised by the ground-truth FP map $\mathbf{M}_{fp}$ and ground-truth FN map $\mathbf{M}_{fn}$ that can be obtained from the ground truth mask $\mathbf{M}$ and the segmentation probability map $\mathbf{P}$ in the following way:  $\mathbf{M}_{fp} = \lnot{\mathbf{M}} \land {(\mathbf{P} \geq \tau)}$; $\mathbf{M}_{fn}=\mathbf{M} \land {(\mathbf{P} < \tau)}$, where $\tau$ is a probability threshold (which is set to 0.5 by default).
Since training the segmentation error decoder is formulated as two binary segmentation tasks, the error decoders and the segmentation decoder can be trained end-to-end in a multi-task learning manner (See Sec. \ref{sec:loss_function}). 

\begin{figure}[t]
  \centering
  \includegraphics[width=7.0cm, height=5.5cm]{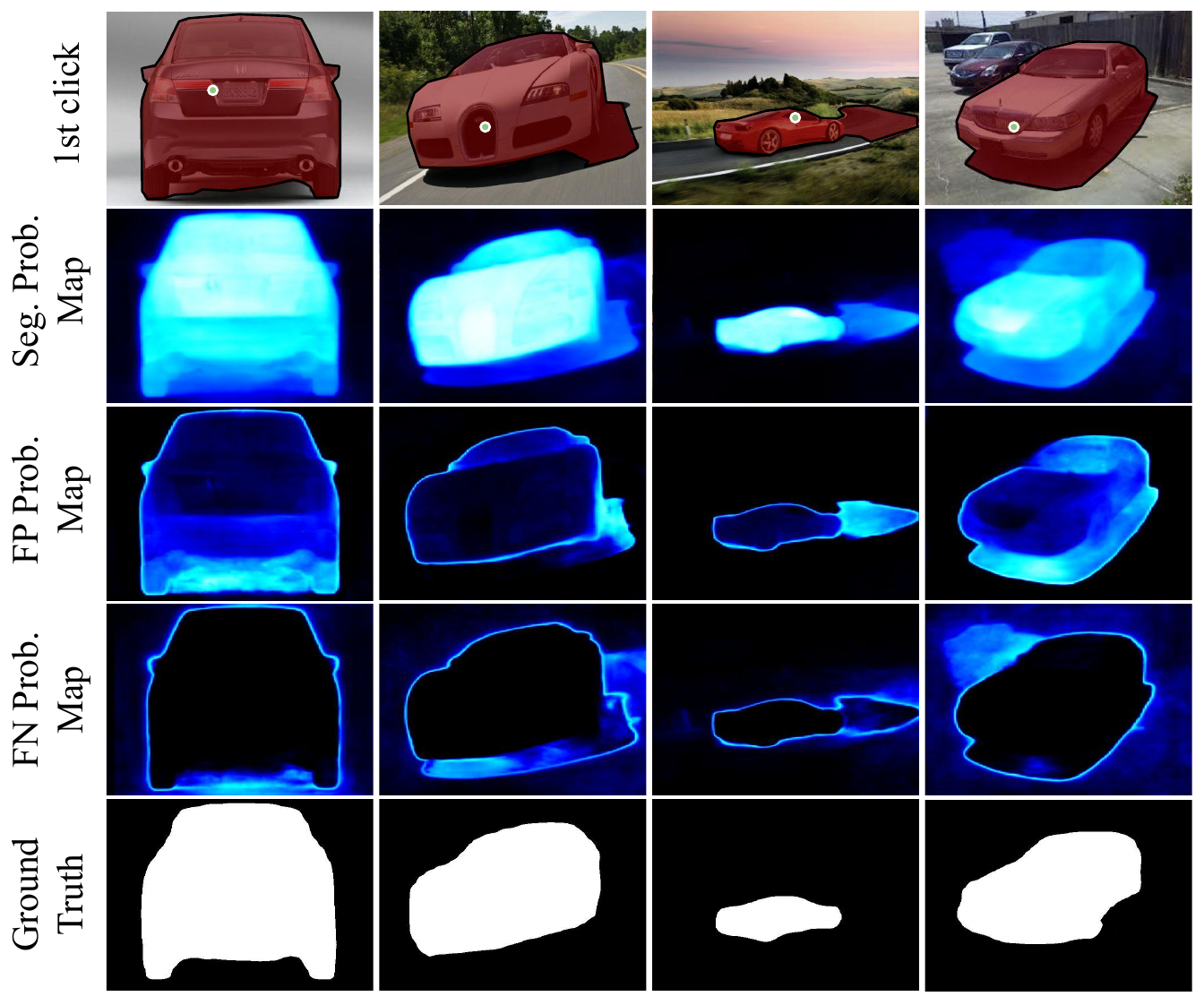}
  \caption{Illustration of the output error maps of {\PseudoClick}. Given the first click (dots in the first row), the network outputs a segmentation map (second row), along with estimated FP and FN error maps (third and fourth rows). Each column represents a test case. All test images are from the Cars dataset~\cite{KrauseStarkDengFei-Fei_3DRR2013}.}
  \label{fig:fp_fn}
\end{figure}

To help readers better understand the function of the proposed error decoder, we show in Fig.~\ref{fig:fp_fn} the error maps generated by the error decoder. We observe that the error maps provide a meaningful estimation of the true errors, as one can easily estimate by comparing the segmentation mask (red masks in first row) with the ground truth (white masks in the last row). The accuracy of these error maps is essential for reliable pseudo clicks generation, introduced next.

\subsection{Pseudo Clicks Generation}
\label{sec:pseudo_clicks_generation}
Given the predicted FP and FN error maps for the current segmentation mask, our method generates one pseudo click from either the FP or the FN map, depending on which map contains the largest error region. First, we transform the two error maps to two binary masks through a predefined threshold (\ie 0.5), followed by extracting a positive/negative click from one of the two binary masks that contains the largest connected error region---the extracted pseudo click locates at the center of this region.
If the largest error region is from the FP mask, then the extracted pseudo click is negative (\ie, indicating that this region should not be segmented); if the largest error region is from the FN mask, then the extracted pseudo click is positive (\ie, indicating that this region should be segmented).
Finally, we encode the new pseudo click as a disk on the encoding maps designed exclusively for pseudo clicks, introduced next.

\subsection{Pseudo Clicks Encoding}
\label{sec:pseudo_clicks_encoding}

As shown in Fig.~\ref{fig:pc_framework}, human clicks and pseudo clicks are encoded separately as small binary disks in the corresponding encoding maps, resulting in two 2-channel disk maps.
Positive clicks are encoded in the positive disk map while the negative clicks are encoded in the negative disk map.
Note that our choice of using disk maps instead of Gaussian maps for clicks encoding is inspired by RITM~\cite{sofiiuk2021reviving}, which shows that disk maps are more efficient and effective than Gaussian maps.
% The radius of the disk for each click is set fixed to 5 pixels, which is purely result-driven.
We perform element-wise addition to merge the feature maps extracted from pseudo clicks and feature maps extracted from the combination of image and human clicks.
The merged feature maps will be fed into the segmentation backbone for end-to-end training.

\subsection{Loss Function}
\label{sec:loss_function}
Although binary cross entropy (BCE) loss is widely used to supervise the interactive segmentation tasks~\cite{zhang2020interactive,jang2019interactive,maninis2018deep,liew2019multiseg,majumder2019content}, we instead use normalized focal loss (NFL)~\cite{sofiiuk2019adaptis}, which allows for faster convergence and better accuracy than BCE, as discussed in \cite{sofiiuk2021reviving}. 
The NFL loss $L$ can be written as:

\begin{equation}
\label{equ:nfl_loss}
\small
\NFL(\mathbf{P}(i, j))= - \frac{1}{\sum\nolimits_{i,j}\mathbf{P}(i, j)}(1 - \mathbf{P}(i,j))^\gamma\log{\mathbf{P}(i,j)}
\end{equation}
where $\mathbf{P}(i,j)$ denotes the prediction $\mathbf{P}$ at point $(i, j)$ and $\gamma>0$ is a tunable focusing parameter (as in the focal loss~\cite{lin2017focal}). 
Since the error branch can be supervised as two binary segmentation tasks (see Sec.~\ref{sec:segmentation_error_decoder}), we also use the NFL loss for them during training. Therefore, the overall loss is a combination of three NFL loss functions:
\begin{equation}
\label{equ:loss}
\small
\begin{split}
\NFL = \sum_{i, j} (\lambda_1 \NFL_{seg}(i, j) + \lambda_2 \NFL_{fp}(i, j) + \lambda_3 \NFL_{fn}(i, j))
\end{split}
\end{equation}
where $\lambda_1, \lambda_2, \lambda_3>0$ represent the weights for each component; $L_{seg}(i, j)$, $L_{fp}(i, j)$, and $L_{fn}(i, j)$ denote $\NFL(\mathbf{P}(i,j))$, $\NFL(\mathbf{E}_{fp}(i, j))$, and $\NFL(\mathbf{E}_{fn}(i,j))$, respectively. 

\subsection{Implementation Details}
\label{sec:implementation_details}

\subsubsection{Click simulation for training and evaluation.}
We automatically simulate human clicks based on the ground truth and current segmentation for fast training and evaluation.
For training, we use a combination of random and iterative click simulation strategies, similar to~\cite{sofiiuk2021reviving}. 
The random click simulation strategy generates a set of positive and negative clicks without considering the order between them~\cite{xu2016deep,sofiiuk2020f,benenson2019large}. 
In contrast, the iterative simulation strategy generates clicks sequentially---a new click is generated based on the erroneous region of a prediction produced by a model using the set of previous clicks~\cite{majumder2019content,kontogianni2020continuous,mahadevan2018iteratively}.
% This process resembles the interaction with a real user. 
% However, a full iterative simulation is very computationally expensive. 
% Instead, we use a combination of the random and iterative simulation strategies to train our iterative models. First, we simulate user clicks with the random simulation strategy. 
% Then, we add from 0 to $N_{iters}$ iteratively simulated clicks.
% We set $N_{iters}$ to 3 based on the results from \cite{sofiiuk2021reviving}.
Once the model is trained, there are two modes to evaluate it: automatic evaluation and human evaluation.
For automatic evaluation, we adopt the iterative click simulation strategy.
Note that the automatically simulated clicks may be different from clicks generated by human evaluation.
We present in the supplementary materials some qualitative results obtained via human evaluation.
% For both training and evaluation, we only generate one pseudo click for each human click since the error of pseudo clicks would soon accumulate without human intervention.

\subsubsection{Previous segmentation as an additional input channel.}
It is natural to incorporate the output segmentation mask from previous interaction as an input for the next correction, providing additional prior information that can help improve the segmentation quality. The previous segmentation mask is added as an additional channel to the RGB image, resulting in a 4-channel image as the input. This 4-channel image  will be concatenated with the human clicks encoding maps, which have two channels (positive and negative clicks are encoded in separate channels). Note that the additional mask input is not shown in Fig.~\ref{fig:pc_framework} for brevity. For the first interaction, we feed an empty mask to our model.

\subsubsection{Post-processing using error maps.}
% The first click should tell the model which object to seg-
% ment, which currently can only be done by the user. Given
% the first click, our model will generate the initial segmen-
% tation as well as the FP&FN errors used for generating the
% first pseudo-click.
Pseudo clicks are extracted from the error maps.
Actually, the error maps can be directly used to refine the segmentation mask (\eg via simple post-processing introduced in Sec.~\ref{sec:comparison_study}).
We argue that post-processing based on FP\&FN maps can be regarded as a by-product of our core contribution---\PseudoClick--–which takes a step further towards the general idea of click imitation.

\section{Experiments}
\label{sec:experiments}

\subsection{Evaluation Details}
\label{sec:evaluation_details}

\subsubsection{Datasets.} 
We evaluate {\PseudoClick} on \textbf{10} public datasets: GrabCut~\cite{rother2004grabcut}, Berkeley~\cite{martin2001berkeley}, DAVIS~\cite{perazzi2016davis}, Pascal~\cite{everingham2010Pascal}, Semantic Boundaries Dataset (SBD) \cite{hariharan2011semantic}, Brain Tumor Segmentation challenge (BraTS)~\cite{baid2021rsna}, ssTEM~\cite{gerhard2013segmented}, Cars~\cite{KrauseStarkDengFei-Fei_3DRR2013}, COCO \cite{lin2014microsoft}, and LVIS \cite{gupta2019lvis}.
Since COCO and LVIS datasets share the same set of images and are complementary to each other in terms of annotation quality and object categories, they can be combined as an ideal training set for the interactive image segmentation task~\cite{sofiiuk2021reviving}. Therefore, we use the combined COCO+LVIS for training and the remaining 8 datasets for evaluation. Specifically, we use the training set of the  COCO+LVIS dataset for training and its validation set for model selection. The Cars~\cite{KrauseStarkDengFei-Fei_3DRR2013} dataset is only used for qualitative evaluation. We test the trained {\PseudoClick} models on the test set of the remaining 7 datasets; no finetuning is conducted on these datasets. For the DAVIS and BraTS datasets, we do not use the original videos or volumes. Instead, we extract 345 and 369 2D slices from the two 3D datasets, respectively. We extracted from each volume in the BraTS the slice that contains the largest tumor area. The two medical image datasets, BraTS~\cite{baid2021rsna} and ssTEM~\cite{gerhard2013segmented}, are used for cross-domain evaluation (see Sec.~\ref{sec:cross_domain_evaluation}) because our models are trained with natural images, which are significantly different from images from medical domain. We refer the readers to the supplementary material for more details.

\begin{figure}[ht]
    \centering
    \includegraphics[width=6.0cm, height=4.2cm, trim=40 5 40 5, clip]{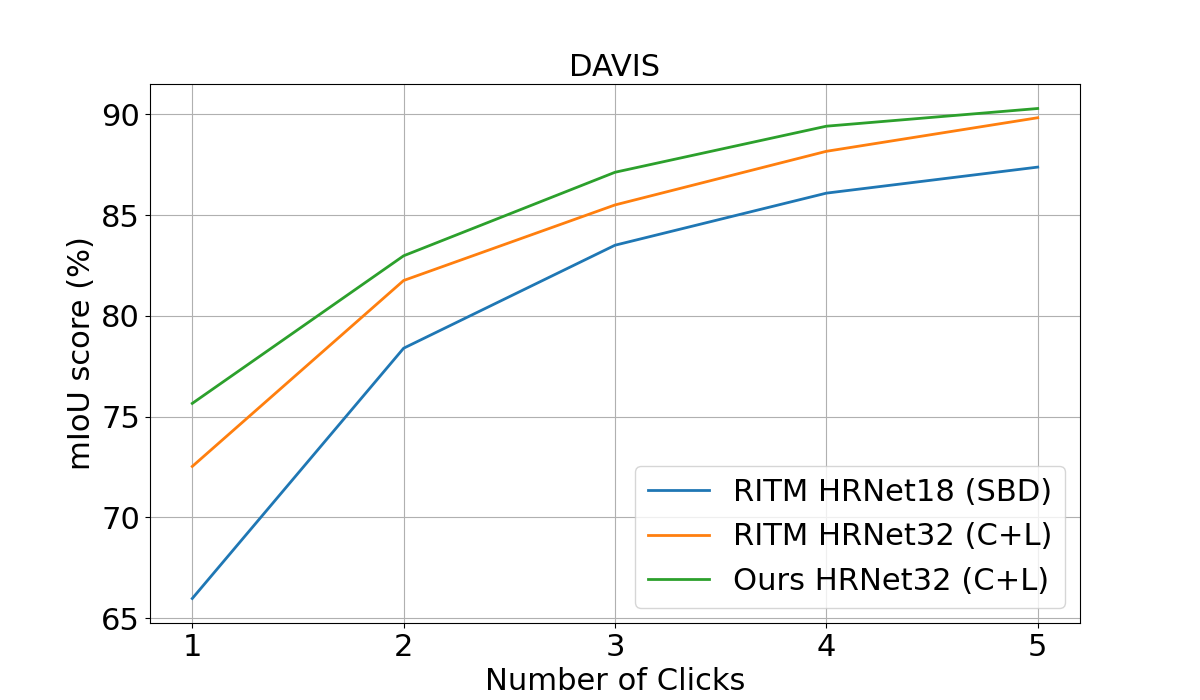}
    \includegraphics[width=6.0cm, height=4.2cm, trim=40 5 40 5, clip]{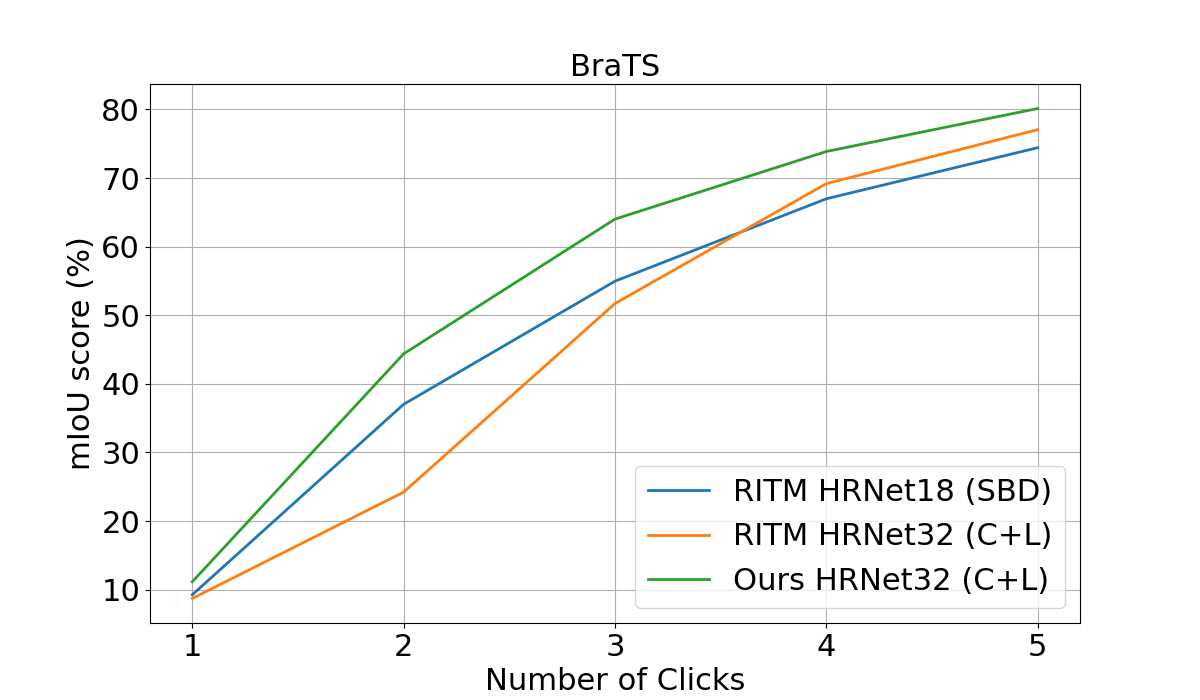}
    \caption{Comparison of mIoU for the first five clicks on DAVIS and BraTS datasets.}
    \label{fig:plots}
\end{figure}

\subsubsection{Segmentation backbone.} 
We choose HRNet-18 and HRNet-32~\cite{wang2020hrnet} as backbones for our {\PseudoClick} model. To show that {\PseudoClick} is a generic framework that can be built upon most of existing segmentation backbones, we also implement {\PseudoClick} on two recently proposed segmentation transformers: SegFormer-B5~\cite{xie2021segformer} and HRFormer-base~\cite{yuan2021hrformer}.
While the two transformers have shown promising preliminary results in our experiments, HRNet still outperforms them to a large margin under the same experimental settings. 

\subsubsection{Evaluation modes and metrics.} 
We evaluate the trained models with two modes: automatic evaluation and human evaluation. For automatic evaluation, we simulate human clicks based on the ground truth and current segmentation mask (see Sec.~\ref{sec:implementation_details}); for human evaluation, a human-in-the-loop will provide clicks based on his/her subjective evaluation.
We use the standard Number of Clicks (NoC) metric that measures the number of clicks required to achieve predefined Intersection over Union (IoU). Specifically, we use NoC@85 and NoC@90 as two main metrics to measure the number of clicks required to obtain 85\% and 90\% IoU, respectively.

\subsubsection{More implementation details.}
We adopt the same data augmentation techniques as in RITM~\cite{sofiiuk2021reviving} for fair comparison.
Pseudo-clicks were not counted in the NoC@85\% or NoC@90\% metrics as they are “free” in terms of human labor.
We implement our models in Python, using PyTorch~\cite{paszke2019pytorch}. We train the models for 200 epochs on COCO+LVIS dataset with an initial learning rate $5\times10^{-4}$, which will decrease to $5\times10^{-5}$ after the epoch 50.
During training, we crop images with a fixed size of $320\times480$ and set the batch size to 32.
We optimize using Adam with $\beta_1=0.9$, $\beta_2=0.999$. 
All models are trained and tested on a single NVIDIA RTX A6000 GPU. 

\begin{table}
\footnotesize
\begin{tabular}{l c c c c c c c}
    \toprule
    \multirow{2}{*}{Method} & GrabCut & Berkeley & SBD & \multicolumn{2}{c}{DAVIS} & \multicolumn{2}{c}{Pascal} \\
    & NoC@90 & NoC@90 & NoC@90 & NoC@85 & NoC@90 & NoC@85 & NoC@90 \\
    \midrule
    GC \cite{boykov2001interactive} $_\text{ICCV01}$
    & 10.00 & 14.22 & 15.96 & 15.13 & 17.41 & - & - \\
    RW \cite{grady2006random} $_\text{PAMI06}$
    & 13.77 & 14.02 & 15.04 & 16.71 & 18.31 & - & - \\
    GM \cite{bai2007geodesic} $_\text{IJCV09}$
    & 14.57 & 15.96 & 17.60 & 18.59 & 19.50 & - & - \\
    GSC \cite{gulshan2010geodesic} $_\text{CVPR10}$
    & 9.12 & 12.57 & 15.31 & 15.35 & 17.52 & - & - \\
    ESC \cite{gulshan2010geodesic} $_\text{CVPR10}$
    & 9.20 & 12.11 & 14.86 & 15.41 & 17.70 & - & - \\
    DIOS \cite{xu2016deep} $_\text{CVPR16}$
    & 6.04 & 8.65 & - & - & 12.58 & 6.88 & - \\
    LD \cite{li2018interactive} $_\text{CVPR18}$
    & 4.79 & - & 10.78 & 5.05 & 9.57 & - & - \\
    BRS  \cite{jang2019interactive} $_\text{CVPR19}$  
    & 3.60 & 5.08 & 9.78 & 5.58 & 8.24 & - & - \\
    f-BRS  \cite{sofiiuk2020f} $_\text{CVPR20}$
    & 2.72 & 4.57 & 7.73 & 5.04 & 7.41 & - & - \\
    IA+SA \cite{kontogianni2020continuous} $_\text{ECCV20}$ 
    & 3.07 & 4.94 & - & 5.16 & - & 3.18 & - \\    
    FCA \cite{lin2020interactive} $_\text{CVPR20}$ 
    & 2.08 & 3.92 & - & - & 7.57 & 2.69 & - \\
    \midrule
    SBD RITM-H18
    & 2.04 & 3.22 & \underline{5.43} & 4.94 & 6.71 & 2.51 & 3.03 \\  
    C+L RITM-H32
    & 1.56 & \underline{2.10} & 5.71 & 4.11 & 5.34 
    & 2.19 & 2.57 \\    
    \midrule
    \midrule
    SBD H18 $_\text{w/ PC}$
    & 2.04 & 3.23 & \textbf{5.40} & 4.81 & 6.57 & 2.34 & 2.74 \\
    SBD H32 $_\text{w/ PC}$
    & 1.84 & 2.98 & 5.61 & 4.74 & 6.16 & 2.37 & 2.78 \\
    C+L H32 $_\text{w/o PC}$
    & \underline{1.55} & 2.11 & 5.68 & \underline{4.09} & \underline{5.27} & \underline{2.14} & \underline{2.52} \\
    C+L H32 $_\text{w/ PC}$
    & \textbf{1.50} & \textbf{2.08} & 5.54 & \textbf{3.79} & \textbf{5.11} & \textbf{1.94} & \textbf{2.25} \\
    \bottomrule
\end{tabular}
\caption{Evaluation results on GrabCut, Berkeley, DAVIS, SBD, and Pascal datasets. Our models are trained on either SBD or COCO+LVIS (denoted as C+L above) datasets. The best results are set in bold; the second best results are shown underlined. ``H18" and ``H32" represent ``HRNet-18" and ``HRNet-32", respectively. ``PC'' means the model is implemented with pseudo clicks. The metrics are NoC@85\% and NoC@90\%, representing the number of clicks required to achieve 85\% and 90\% IoU, respectively.}
\label{tab:res_main}
\end{table}

\begin{figure}
    \centering
    \includegraphics[width=\linewidth]{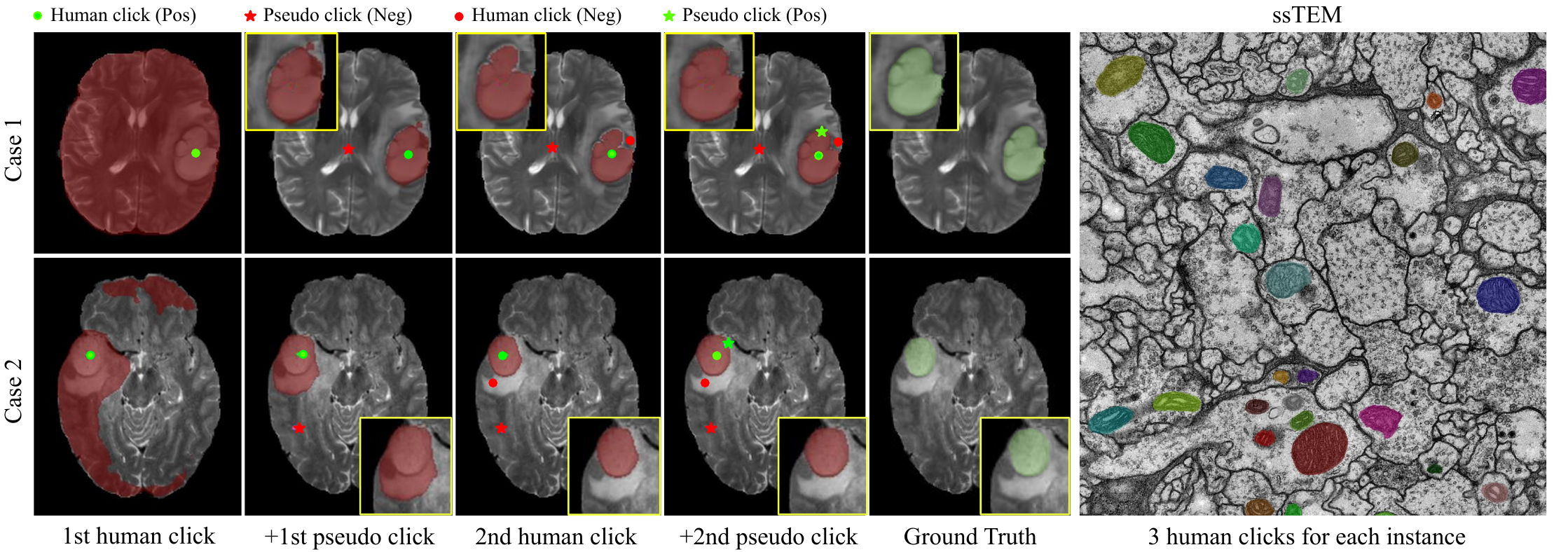}
    \caption{Cross-domain evaluation on two medical image datasets: BraTS~\cite{baid2021rsna} (left) and ssTEM~\cite{gerhard2013segmented} (right). The evaluation is performed by a human annotator through our internally developed interactive segmentation GUI. For the BraTS dataset, we use two clicks for each image. For the ssTEM dataset, we strictly use three clicks for each instance. Note that our model is trained on natural images, but shows very robust results on the two medical datasets.}
    \label{fig:qualitative_brats}
\end{figure}

\subsection{Comparison with State-of-the-Art}
\label{sec:comparison_with_sota}
\begin{figure*}[t]
  \centering
  \includegraphics[width=\linewidth, height=4.0cm]{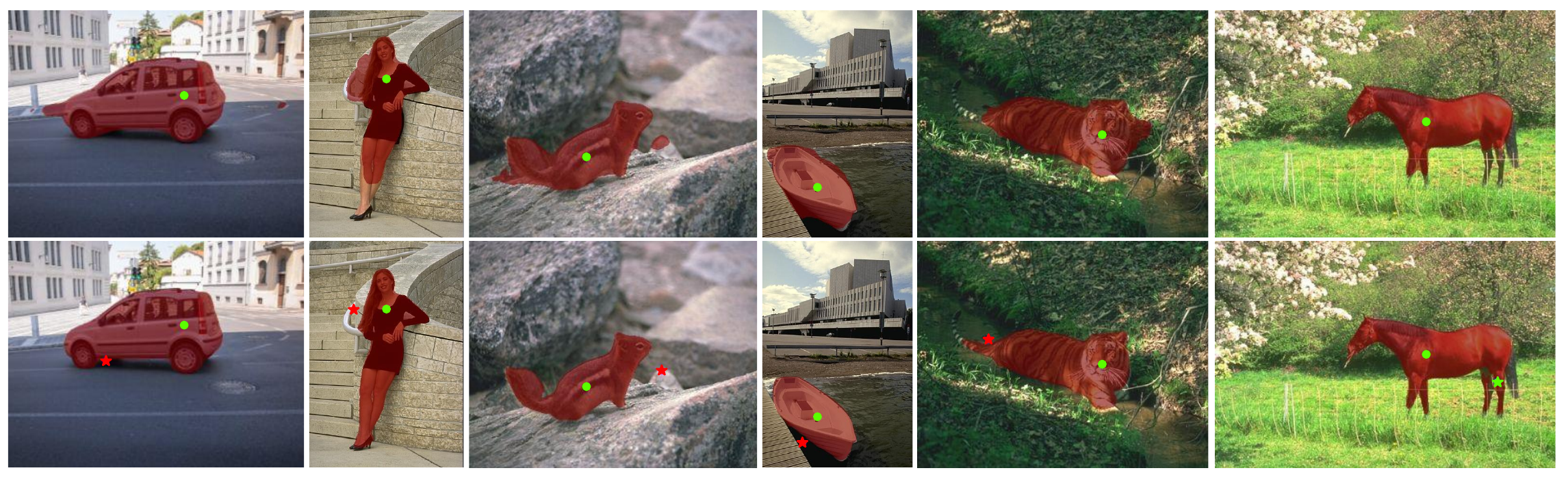}
  \includegraphics[width=\linewidth, height=3.3cm]{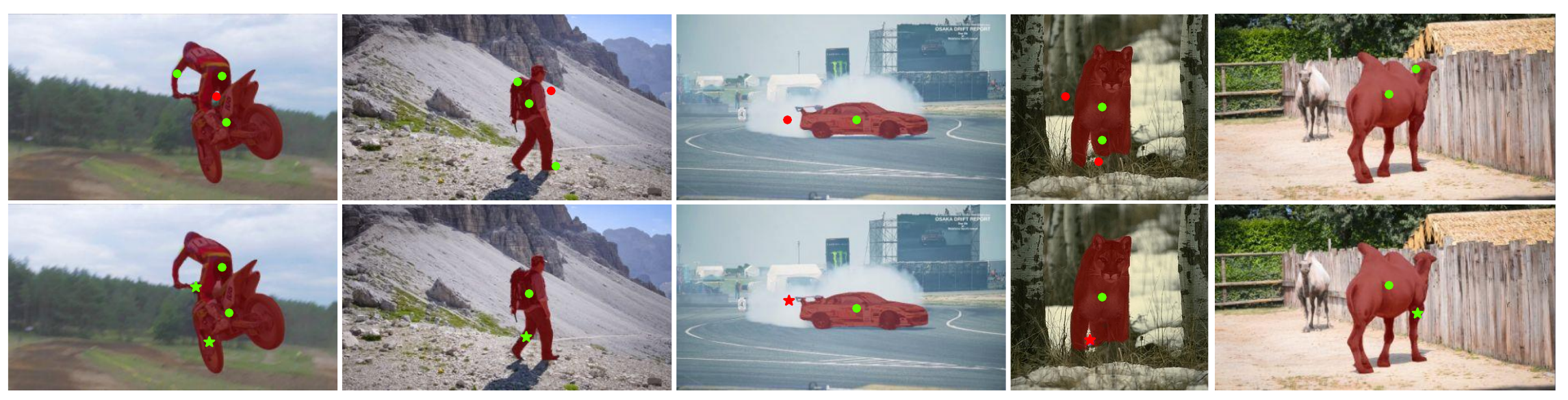}
  \caption{Qualitative evaluation of {\PseudoClick} model on natural images. Top: segmentation results with one pseudo click. The first row shows the results of the first human click; the second row shows the improvement of the segmentation mask with the help of the first pseudo click. Bottom: segmentation results with IoU greater than 90\% given several clicks. The first row shows the segmentation using only human clicks. The second row shows the segmentation with both human and pseudo clicks. The color and shape of a click follow the same rule shown in Fig.~\ref{fig:qualitative_brats}.}
  \label{fig:qualitative}
\end{figure*}

We compare our results with existing state-of-the-art methods, including f-BRS~\cite{sofiiuk2020f}, IA+SA~\cite{kontogianni2020continuous}, FCA~\cite{lin2020interactive}, and RITM~\cite{sofiiuk2021reviving}. Tab.~\ref{tab:res_main} shows the quantitative results.
Our proposed {\PseudoClick} approach outperforms all the methods across the five benchmark datasets. For example, compared with RITM on the Pascal dataset, our model uses 12.4\% and 11.4\% fewer number of clicks for achieving 85\% and 90\% IoU, respectively.
Fig.~\ref{fig:plots} shows plots for mean IoU with respect to the first several clicks for the DAVIS and BraTS datasets. Our approach shows continuous improvement in terms of accuracy and stability.
We also visualize the evaluation process of the proposed method on some images in Fig.~\ref{fig:qualitative}. From the qualitative results, we can see that the pseudo clicks automatically generated from our model can accurately focus on false positive and false negative regions. Hence, they are able to refine the predicted segmentation masks and thereby alleviate human annotation effort.
Computational analysis is shown in Fig.~\ref{tab:ablation_computational_analysis}.

\subsection{Cross-Domain Evaluation}
\label{sec:cross_domain_evaluation}
To evaluate the generalization cability of the proposed method, we conduct cross-domain evaluation on two medical image datasets: BraTS~\cite{baid2021rsna} and ssTEM~\cite{gerhard2013segmented}. Specifically, we directly apply our {\PseudoClick} models trained on SBD or COCO+LVIS datasets to the medical images without finetuning (medical images in grayscale are replicated 3 times channel-wise to be of the same channel dimension with RGB images). We report cross-domain evaluation results in Tab.~\ref{tab:cross_domain_brats} and Tab.~\ref{tab:ablation_ssTEM}. We observe that our models generalize very well to medical images without fine-tuning. 
Note that these results are evaluated by a human annotator. 
Some qualitative results on the two medical datasets are shown in Fig.~\ref{fig:qualitative_brats}.

\begin{table}
\footnotesize
\renewcommand\arraystretch{0.9}
\setlength{\tabcolsep}{0.6mm}{
  \centering
  \begin{tabular}{l c c c c c c}
    \toprule
    Method & Train & Finetune & Backbone & mIoU@3 & mIoU@5 \\
    \midrule
    RITM \cite{sofiiuk2021reviving} & SBD & N/A & HRNet18 & 54.9 & 74.4 \\
    RITM \cite{sofiiuk2021reviving} & C+L & N/A & HRNet32 & 51.7 & 77.1 \\
    \midrule
    Ours & SBD & N/A & HRNet18 & 54.5 & 74.6 \\
    Ours & C+L & N/A & HRNet32 & \textbf{64.0} & \textbf{80.1} \\
    \bottomrule
  \end{tabular}
  \caption{Cross-domain evaluation on the BraTS dataset. The evaluation measure is mean IoU (\%) given 3 or 5 human clicks.}
  \label{tab:cross_domain_brats}
 }
 
\setlength{\tabcolsep}{0.25mm}{
  \centering
  \begin{tabular}{l c c c c c}
    \toprule
    Method & Train & Finetune & Backbone & \#Clicks & mIoU \\
    \midrule
    Curve-GCN~\cite{ling2019fast} & CityScapes & N/A & ResNet-50 & 2 & 60.9 \\
    IOG~\cite{zhang2020interactive} & Pascal & N/A & ResNet-101 & 3 & 83.7 \\
    RITM~\cite{sofiiuk2021reviving} & SBD & N/A & HRNet18 & 3 & 77.3 \\
    RITM~\cite{sofiiuk2021reviving} & C+L & N/A & HRNet32 & 3 & 86.4 \\
    \midrule
    Ours & SBD & N/A & HRNet18 & 3 & 80.9 \\
    Ours & C+L & N/A & HRNet32 & 3 & \textbf{87.2} \\
    \bottomrule
  \end{tabular}
  \caption{Cross-domain evaluation on the ssTEM~\cite{gerhard2013segmented} dataset. The evaluation measure is mean IoU (\%) given 2 or 3 human clicks. The results for IOG and Curve-GCN methods are copied from the corresponding papers.}
  \label{tab:ablation_ssTEM}}
 
\end{table}

% \begin{table}
% \footnotesize
% \renewcommand\arraystretch{0.9}
% \setlength{\tabcolsep}{0.25mm}{
%   \centering
%   \begin{tabular}{l c c c c c}
%     \toprule
%     Method & Train & Finetune & Backbone & \#Clicks & mIoU \\
%     \midrule
%     Curve-GCN~\cite{ling2019fast} & CityScapes & N/A & ResNet-50 & 2 & 60.9 \\
%     IOG~\cite{zhang2020interactive} & Pascal & N/A & ResNet-101 & 3 & 83.7 \\
%     RITM~\cite{sofiiuk2021reviving} & SBD & N/A & HRNet18 & 3 & 77.3 \\
%     RITM~\cite{sofiiuk2021reviving} & C+L & N/A & HRNet32 & 3 & 86.4 \\
%     \midrule
%     Ours & SBD & N/A & HRNet18 & 3 & 80.9 \\
%     Ours & C+L & N/A & HRNet32 & 3 & \textbf{87.2} \\
%     \bottomrule
%   \end{tabular}
%   \caption{Cross-domain evaluation on the ssTEM~\cite{gerhard2013segmented} dataset. The evaluation measure is mean IoU (\%) given 2 or 3 human clicks. The results for IOG and Curve-GCN methods are copied from the corresponding papers.}
%   \label{tab:ablation_ssTEM}}
% \end{table}

\subsection{Comparison Study}
\label{sec:comparison_study}

\subsubsection{Segmentation backbone comparison.}
We have demonstrated in Tab.~\ref{tab:res_main} that our method outperforms existing state-of-the-art when using HRNet-32 as its backbone. In this study, we implement other backbones including two recently proposed vision transformers, SegFormer~\cite{xie2021segformer} and HRFormer~\cite{yuan2021hrformer}, that show encouraging results when compared with CNNs. Tab.~\ref{tab:ablation_backbone} shows the comparison results. 
All models are pre-trained on ImageNet~\cite{deng2009imagenet} and finetuned on COCO+LVIS dataset with NFL loss function. We use SegFormer-B5~\cite{xie2021segformer} and HRFormer-Base~\cite{yuan2021hrformer} for the transformers. Models are trained and evaluated using pseudo clicks. The evaluation measure is NoC@85\%.
While the transformers achieve decent results, we actually spend little time in tuning the hyper-parameters and modifying the architecture when transferring from CNNs to transformers. This demonstrates the flexibility and generalization capability of our framework.

\subsubsection{Loss functions comparison.} In this study, we train {\PseudoClick} models using four different loss functions: binary cross entropy (BCE) loss, focal loss (FL)~\cite{lin2017focal}, Soft IoU loss~\cite{rahman2016optimizing}, and normalized focal loss (NFL)~\cite{sofiiuk2019adaptis}. 
% We test our models on four datasets: Berkeley, SBD, DAVIS, and Pascal. 
Each experiment uses the HRNet32 model. All the four models are trained on the COCO+LVIS dataset. 
Results in Tab.~\ref{tab:ablation_loss} show that training with NFL leads to the best accuracy.

\begin{table}
\footnotesize
\parbox{.45\linewidth}{
\centering
\begin{tabular}{l c c c c}
    \toprule
    Backbone & Berkeley & SBD & DAVIS & Pascal \\
    \midrule
    ResNet51  & 1.94 & 3.49 & 4.87 & 2.18 \\
    ResNet101 & 1.85 & 3.44 & 4.14 & 2.02 \\
    SegFormer & 2.54 & 4.10 & 4.11 & 2.26 \\
    HRFormer  & 1.84 & 4.53 & 4.80 & 2.62 \\
    \bottomrule
\end{tabular}
\caption{Comparison study for different segmentation backbones.}
\label{tab:ablation_backbone}
}
\hfill
\parbox{.45\linewidth} {
\centering
\begin{tabular}{l c c c c}
    \toprule
    Loss & Berkeley & SBD & DAVIS & Pascal \\
    \midrule
    BCE & 1.44 & 3.53 & 3.97 & 1.98 \\
    FL & 1.43 & 3.54 & 3.79 & 2.01 \\
    Soft IoU & 1.44 & 3.63 & 3.96 & 2.10 \\
    NFL & 1.40 & 3.46 & 3.79 & 1.94 \\
    \bottomrule
\end{tabular}
\caption{Comparison study for different loss functions.}
\label{tab:ablation_loss}
}
\end{table}

\begin{table}
\footnotesize
\parbox{.4\linewidth}{
\centering
\begin{tabular}{l c c c c}
  \toprule
  Train & Berkeley & SBD & DAVIS & Pascal \\
  \midrule
  Pascal & 2.33 & 5.87 & 5.67 & 2.66 \\
  SBD & 1.67 & 3.51 & 4.74 & 2.37 \\
  LVIS & 2.63 & 5.40 & 6.97 & 3.14 \\
  C+L & 1.40 & 3.46 & 3.79 & 1.94 \\
  \bottomrule
\end{tabular}
\caption{Comparison study for different training datasets.}
\label{tab:ablation_dataset}
}
\hfill
\parbox{.52\linewidth} {
\centering
\begin{tabular}{l c c c}
  \toprule
  Model & Param/M & FLOPs/G & Speed/s \\
  \midrule
  RITM-H32 & 30.95 & 16.57 & 0.137\\
  Ours-H32-PC & 36.79 & 18.43 & 0.185\\
  \bottomrule
\end{tabular}
\caption{Computational analysis. Speed is measured as second per click (including a pseudo click for ours) with a NVIDIA A6000 GPU.}
\label{tab:ablation_computational_analysis}
}
\end{table}

\subsubsection{Training datasets comparison.} In this study, we train {\PseudoClick} models on four different training datasets: Pascal, SBD, LVIS, and COCO+LVIS. We test on four datasets: Pascal, SBD, Berkeley, and DAVIS. For each experiment, our model is based on HRNet32 and is trained with NFL loss function. We report results in Tab.~\ref{tab:ablation_dataset}. We observe that the model trained on COCO+LVIS shows the best performance, highlighting the benefit of combining COCO and LVIS for training interactive segmentation models. We also notice that on the Pascal dataset, model trained on COCO+LVIS dataset is even better than the model trained on Pascal dataset. This, again, highlights the strengths of COCO+LVIS dataset: 1) large dataset size. The number of annotated instances in COCO+LVIS dataset is $50\times$ and $170\times$ times larger than SBD and Pascal, respectively; 2) diverse and high annotation quality. 

\subsubsection{Post-processing vs. pseudo clicks.} In this study, we directly use the two error maps for refining the segmentation mask. 
The two error maps serve as a regularization for the segmentation branch during training. 
As shown in Fig. \ref{fig:fp_fn}, they quantize the segmentation mask reasonably well, and thus can be used for refining the segmentation mask. 
To achieve this goal, we simply subtract the two error maps from the segmentation map (all three maps are probability maps). 
We compare the post-processing with adding one pseudo click. 
The comparison results are shown in Tab.~\ref{tab:post_processing}. 
We emphasize that the post-processing based on FP\&FN maps can also be regarded as a contribution of our work as it is a by-product of our core contribution. 
Given two human clicks on the BraTS dataset, the relative mIoU obtained by adding one pseudo-click is 5.2\% higher than the mIoU by post-processing, which is substantial considering the strong performance of post-processing.

\begin{table}
\footnotesize
\renewcommand\arraystretch{0.9}
\setlength{\tabcolsep}{2.0mm}{

    \centering
    \begin{tabular}{l c c c c c c}
    \toprule
    & \multicolumn{3}{c}{BraTS} & \multicolumn{3}{c}{DAVIS} \\
    \cmidrule(lr){2-4} \cmidrule(lr){5-7}
    mIoU@Human-clks & 2 & 3 & 5 & 2 & 3 & 5 \\
    \midrule
    Baseline (BL) & 23.2 & 51.2 & 77.3 & 80.2 & 85.6 & 89.3 \\
    BL+post-processing & 42.6 & 63.5 & 79.7 & 81.3 & 86.2 & 90.1 \\
    BL+1 pseudo-click & \textbf{44.8} & \textbf{64.0} & \textbf{80.1} & \textbf{83.7} & \textbf{87.4} & \textbf{90.8} \\
    \bottomrule
    \end{tabular}
    \caption{Adding one pseudo-click vs.\ post-processing. Comparison of post-processing and pseudo clicks for mask refinement given a fixed number of human clicks. The Baseline model above is our best {\PseudoClick} model (C+L HRNet32) on the BraTS dataset.}
    \label{tab:post_processing}
 }
\end{table}

% \begin{table}
% \footnotesize
% \centering
% \begin{tabular}{l c c c c}
%     \toprule
%     Backbone & Berkeley & SBD & DAVIS & Pascal \\
%     \midrule
%     ResNet51  & 1.94 & 3.49 & 4.87 & 2.18 \\
%     ResNet101 & 1.85 & 3.44 & 4.14 & 2.02 \\
%     SegFormer & 2.54 & 4.10 & 4.11 & 2.26 \\
%     HRFormer  & 1.84 & 4.53 & 4.80 & 2.62 \\
%     \bottomrule
% \end{tabular}
% \caption{Comparison study for segmentation backbones. All models are pre-trained on ImageNet~\cite{deng2009imagenet} and finetuned on COCO+LVIS dataset with NFL loss function. We use SegFormer-B5~\cite{xie2021segformer} and HRFormer-Base~\cite{yuan2021hrformer} for this experiment. All models are trained and evaluated using pseudo clicks. The evaluation measure is NoC@85\%.}
% \label{tab:ablation_backbone}
% \end{table}

% \begin{table}
% \footnotesize
% \renewcommand\arraystretch{0.9}
%   \centering
%   \begin{tabular}{l c c c c}
%     \toprule
%     Loss & Berkeley & SBD & DAVIS & Pascal \\
%     \midrule
%     BCE & 1.44 & 3.53 & 3.97 & 1.98 \\
%     FL & 1.43 & 3.54 & 3.79 & 2.01 \\
%     Soft IoU & 1.44 & 3.63 & 3.96 & 2.10 \\
%     NFL & 1.40 & 3.46 & 3.79 & 1.94 \\
%     \bottomrule
%   \end{tabular}
%   \caption{Comparison study for four loss functions: binary cross entropy (BCE), focal loss~\cite{lin2017focal} (FL), Soft IoU loss~\cite{rahman2016optimizing}, and normalized focal loss (NFL). The {\PseudoClick} model is based on HRNet32 and is trained on the COCO+LVIS dataset. The evaluation measure is NoC@85\%.}
%   \label{tab:ablation_loss}
% \end{table}

\section{Limitations}

The major limitation of the proposed method is that pseudo clicks may not be as accurate as human clicks, and therefore may cause the segmentation accuracy to drop. This may lead to extra work for users to withdraw the poorly placed pseudo clicks or to correct the error by putting more points. Fortunately, this issue has be greatly alleviated by separating the encoding maps for the two types of clicks. By separating the two encoding maps, the inaccurate pseudo clicks are tolerated during the training and less likely to cause accuracy to drop during evaluation. In the early stage of this project, we discovered this issue. After separating the two types of encoding maps, as implemented in our current architecture, this issue has been significantly eliminated. 
% Another weakness of our method is that it can not predict the first click, meaning the user has to click first. 
% The first click should tell the model which object to segment, which currently can only be done by the user. 
% Given the first click, our model will generate the initial segmentation as well as the FP\&FN errors used for generating the first pseudo-click.
% Since first click generation is a much more challenging task than next-click generation, we leave it for future work.

%------------------------------------------------------------------------
\section{Conclusion}
\label{sec:conclusion}
We proposed {\PseudoClick}, a novel interactive segmentation framework that
automatically imitates human clicks and efficiently segments objects with the imitated pseudo clicks. {\PseudoClick} is a general framework that can be built upon different types of segmentation backbones, including both CNNs and transformers, with little effort in tuning the hyper-parameters and modifying the network architectures. 
We evaluated {\PseudoClick} thoroughly on benchmarks from multiple domains and modalities with extensive comparison and cross-domain evaluation experiments that demonstrated the effectiveness as well as the generalization capability of the proposed method.

\section*{Acknowledgements}
Research reported in this publication was supported by the National Institutes of Health (NIH) under award number NIH 1R01AR072013. 
The content is solely the responsibility of the authors and does not necessarily represent the official views of the NIH.
% \clearpage
\bibliographystyle{ieeetr}
\bibliography{eccv22}
\end{document}